\documentclass[runningheads]{llncs}
\usepackage[T1]{fontenc}
\usepackage{graphicx}
\usepackage{amsmath}
\usepackage{amssymb}

\usepackage{amsthm}
\usepackage{graphicx}
\usepackage{subcaption}
\usepackage{xcolor}
\usepackage{stmaryrd}
\usepackage[noend,ruled,vlined,linesnumbered]{algorithm2e}
\usepackage{comment}
\usepackage[font=small,labelfont=bf]{caption}
\usepackage{url}
\usepackage{cite}

\newcommand{\mD}{\mathcal{D}}

\newcommand{\mT}{\mathcal{T}}
\newcommand{\mU}{\mathcal{U}}
\newcommand{\mX}{\mathcal{X}}
\newcommand{\Exp}{\mathbb{E}}
\newcommand{\R}{\mathbb{R}}
\newcommand{\be}{\mathbf{e}}

\newcommand{\by}{\mathbf{y}}
\newcommand{\tp}{\mathsf{T}}

\newcommand{\abs}[1]{\left\vert #1 \right\vert}
\newcommand{\bbracket}[1]{\left\llbracket #1 \right\rrbracket}

\SetKw{KwInit}{Initialize:}
\SetKw{KwIn}{Input:}
\SetKw{KwOut}{Output:}

\begin{document}
\title{Stackelberg Meta-Learning Based Shared Control for Assistive Driving
\thanks{This work has been submitted to the IROS 2024 for review.}
}
%
%\titlerunning{Stackelberg Meta-Learning Based Shared Control}
\author{Yuhan Zhao\inst{1} \and
Quanyan Zhu\inst{1}}
\authorrunning{Y. Zhao and Q. Zhu}
\institute{New York University, Brooklyn NY 11201. \\
\email{\{yhzhao, qz494\}@nyu.edu}}
\maketitle              % typeset the header of the contribution
\begin{abstract}
Shared control allows the human driver to collaborate with an assistive driving system while retaining the ability to make decisions and take control if necessary. However, human-vehicle teaming and planning are challenging due to environmental uncertainties, the human's bounded rationality, and the variability in human behaviors. An effective collaboration plan needs to learn and adapt to these uncertainties.
To this end, we develop a Stackelberg meta-learning algorithm to create automated learning-based planning for shared control. The Stackelberg games are used to capture the leader-follower structure in the asymmetric interactions between the human driver and the assistive driving system. The meta-learning algorithm generates a common behavioral model, which is capable of fast adaptation using a small amount of driving data to assist optimal decision-making. 
We use a case study of an obstacle avoidance driving scenario to corroborate that the adapted human behavioral model can successfully assist the human driver in reaching the target destination. Besides, it saves driving time compared with a driver-only scheme and is also robust to drivers' bounded rationality and errors\footnote{The simulation codes are available at \url{https://github.com/yuhan16/Stackelberg-Assistive-Driving}.}.

%\keywords{First keyword  \and Second keyword \and Another keyword.}
\end{abstract}
\section{Introduction} \label{sec:intro}
The increasing affordability of robots and related technologies has made human-robot teaming more accessible. The coordination and collaboration between humans and robots revolutionize the way work is performed and help improve efficiency and performance in various domains, such as collective transportation \cite{rozo2015learning,yu2021adaptive} and manufacturing \cite{matheson2019human,wang2020overview}. Shared control is one of the essential teaming mechanisms in autonomous driving \cite{wang2020decision,marcano2020review,xing2021toward}. It augments human drivers with an advanced driver assistance system (ADAS) to enhance safety, comfort, and efficiency while retaining the ability of the human driver to make decisions and take control if necessary. 

Shared control provides a convenient scheme for human participation and interventions. However, several challenges arise with implementing shared control in driving.
First, human drivers have limited cognitive capacity to process information, resulting in limited computational and planning capabilities compared with onboard computers. It also leads to asymmetric collaboration between the ADAS and the human driver because the ADAS is required to take more planning responsibilities to take care of the driver. There is a need for asymmetric collaboration frameworks to deal with it.
Second, human driver behavioral models (e.g., utility function) are associated with uncertainties and bounded rationality. Learning becomes an essential tool to estimate the driver model since it is hard to obtain in practice.

\begin{figure}[t]
    \centering
    \includegraphics[height=5cm]{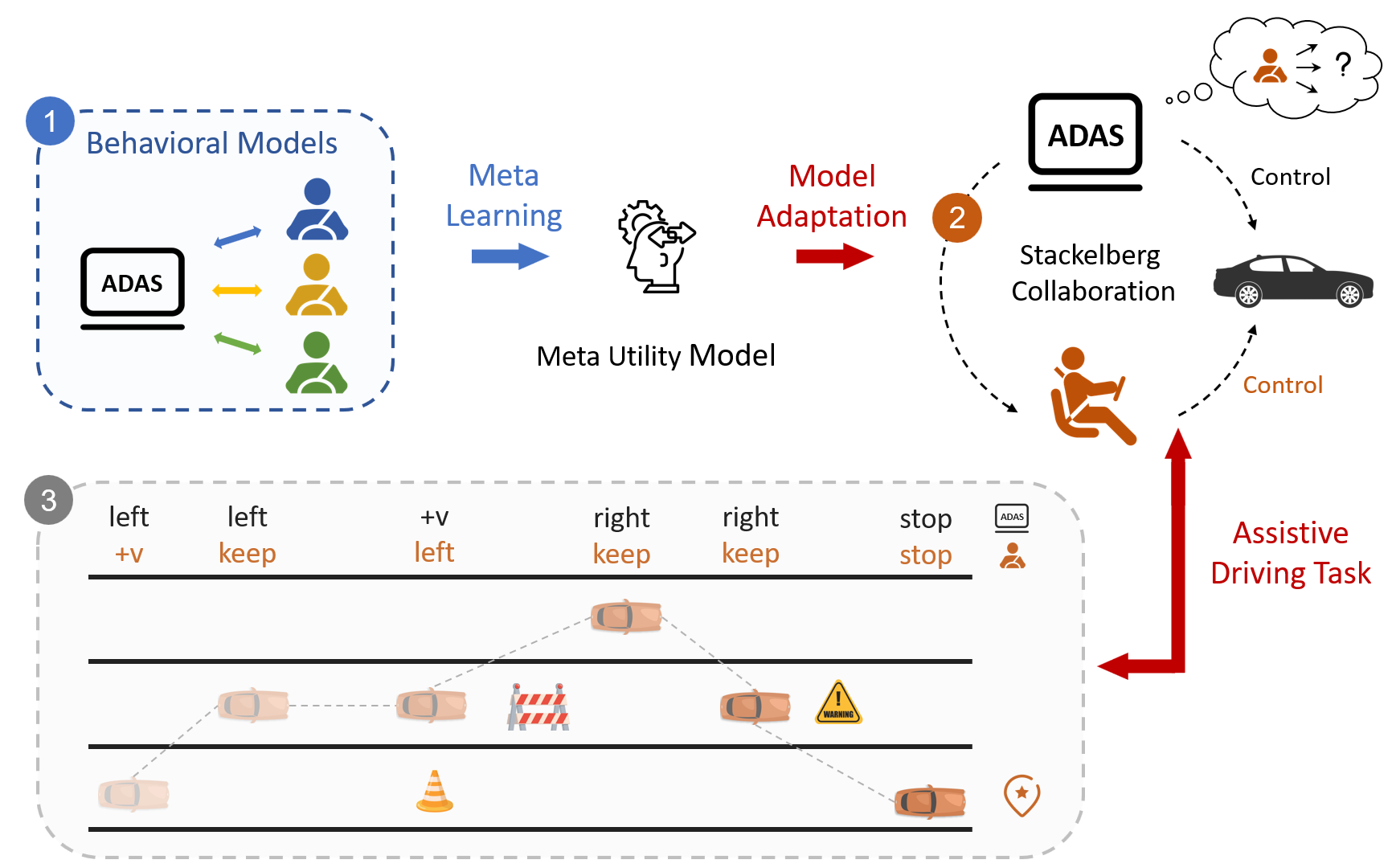}
    \caption{Illustration of Stackelberg shared control framework. The ADAS leverages meta-learning (1) to compute a meta utility model for different human drivers and adapts it to a driver-specific model to perform Stackelberg collaboration (2) for assistive driving tasks. We verify the algorithms using an obstacle avoidance driving scenario (3).}
    \label{fig:intro}
\end{figure}

To establish the asymmetric collaboration framework, we model the interaction between the human driver and the ADAS as a dynamic Stackelberg game \cite{bacsar1998dynamic,simaan1973stackelberg}. Stackelberg games provide a quantitative framework to characterize the leader-follower type of asymmetric interactions and have been applied in many domains in robotics, such as autonomous driving \cite{sadigh2016planning,fisac2019hierarchical} and human-robot interaction \cite{nikolaidis2017game,tian2022safety}. In our work, the ADAS, acting as the leader, searches over the set of admissible trajectories by predicting the human driver's behavior and selects the ones that are most beneficial to the driver. The human driver acts as the follower and responds to ADAS’s strategy. An illustrative diagram of the framework is shown in Fig.~\ref{fig:intro}.

A learning-based quantal response (QR) model \cite{mckelvey1995quantal} is consolidated into the Stackelberg shared control framework to deal with uncertainties in human drivers' decision-making. The QR model uses the logit choice model to abstract the agent’s probabilistic choice of actions. It is a well-validated model to characterize the human’s decision on the noisy perception of her utility under cognitive limitations, and it has been successfully used in behavioral economics \cite{rogers2009heterogeneous,mckelvey1998quantal} and robotics \cite{stefansson2019human,tian2021anytime}.

Another essential challenge for learning arises from the need to adapt to variabilities in human behaviors. Meta-learning is such a scheme that learns a customizable plan for solving a specific task from a prescribed set of tasks \cite{finn2017model,hospedales2021meta}. Specifically, it is used to learn a generalized decision-making model (utility function) of different types of human drivers and adapt to a specific driver only using a small amount of learning data. Based on the bespoke model, the ADAS makes effective planning strategies.

In this work, we develop a meta-learning approach to address diverse driver-vehicle interactions under the Stackelberg shared control framework. In particular, a vehicle faces a family of human drivers with varied driving behaviors (or types). The driving pattern of each human driver is characterized by her utility function. Once a specific driver requests to drive the vehicle, the ADAS can quickly generate a driver-specific model based on the average utility and perform effective driving assistance using the adapted driver model. We create a three-lane obstacle avoidance driving scenario (block 3 in Fig.~\ref{fig:intro}) to evaluate the developed algorithms. The simulation results show that the ADAS can successfully assist different types of human drivers in reaching the target destination compared with the driver-only scheme. Besides, it also shows that the ADAS with meta-learned and adapted driver model is also robust to the decision-making uncertainty caused by the human driver's bounded rationality and errors. 

\emph{Notations:} We use superscript $L$ and $F$ to denote the leader and the follower-related quantities, respectively. We use $\bbracket{n}$ to represent the set $\{0,1,\dots, n\}$.

\section{Related Work}
Driver-vehicle interactions have been studied in the literature by many approaches, such as rule-based methods \cite{li2018shared,saito2018control} and game theory \cite{li2015continuous,na2014game}. Extensive literature implicitly assumes that human drivers have the same decision-making speed as onboard controllers \cite{flad2017cooperative}, or assume that human drivers play the same-level role in planning as onboard controllers \cite{hang2020human,li2022indirect}. For example, the differential game theoretic model for cooperative driving developed by Flad et al. in \cite{flad2017cooperative} assumes that the driver responds to the vehicle system in a continuous time fashion. They are yet sufficient for the asymmetric interactions induced by human cognitive limitations. Fisac et al. in \cite{fisac2019hierarchical} has proposed a hierarchical game-theoretic framework for vehicle trajectory planning, but the human driver needs to make decisions at every decision step. %Our work further relaxes this requirement and develops new planning algorithms.

The use of the QR model has a long history in modeling human behavior in robotics and autonomous driving. Recent studies have begun to focus on learning-based approaches to estimate the QR model. For example, Wu et al. in \cite{wu2022inverse} has developed a learning algorithm to recover the follower’s utility in a repeated static Stackelberg game based on the follower's quantal response. In our work, we generalize the learning approach into dynamic Stackelberg games and meta-learning contexts to design assistive and fast-adapted driving strategies for different human drivers. 

Meta-learning has been used as a promising learning approach to adapt to different tasks in robotics. For example, Xu and Zhu in \cite{xu2022meta} have developed meta-learning and adaptation algorithms to find fast policy-centric motion planners for a class of motion planning problems. Richards et al. in \cite{richards2021adaptive} has leveraged meta-learning to design adaptive controllers for nonlinear systems to adapt to environment uncertainty.

\section{Problem Formulation} \label{sec:problem}
\subsection{Shared Control as Dynamic Stackelberg Games}
We formulate the shared control between a planner\footnote{For simplicity, we use the ``planner" to refer to the ADAS in the following.} and a human driver as a dynamic Stackelberg game.
We denote $x \in \mX$ as the vehicle's state, including the position, lane number, and velocity. Let $u^L \in \mU^L$ and $u^F \in \mU^F$ be the planner and the human driver's actions and action sets. A special action $\varnothing \in \mU^L$ and $\varnothing \in \mU^F$ mean that the planner and the driver keep the current vehicle status and take no actions. Let $f: \mX \times \mU^L \times \mU^F \to \mX$ be the transition dynamics and $g^L: \mX \times \mU^L \times \mU^F \to \R$ (resp. $g^F$) be the planner's (resp. human driver's) utility function. We consider discrete states and actions for high-level driving strategy planning. The dimensions of the state and action sets are given by $\abs{\mX} = n$, $\abs{\mU^L} = m^L$, and $\abs{\mU^F} = m^F$. We define leader's mixed strategy $y^L \in \Delta(m^L)$ (resp. $y^F \in \Delta(m^F)$) over the simplex set $\Delta(m^L) := \{ y \in \R^{m^L} \vert \sum_{u^L} y(u^L) = 1, y \geq 0 \}$, where $y^L(u^L)$ is the probability of choosing the action $u^L \in \mU^L$.

We set the decision horizon in the Stackelberg game as $T$ and use the feedback Stackelberg equilibrium (FSE) to characterize the collaborative driving plan, meaning that the planner and the human driver play the FSE strategy to drive the vehicle.
Due to limited cognitive capacity, human drivers may not perform as fast decision-making as onboard computers. We assume that the human driver only makes decisions at some interaction stages. We define a decision indicator function $\sigma(t)$ such that $\sigma(t) = 1$ means that the human driver makes planning decisions at time $t$, while $\sigma(t) = 0$ means the driver does not make decisions and the vehicle is fully controlled by the planner. We have $\sum_{t=0}^{T-1} \sigma(t) = K < T$. The planner consistently makes decisions at each time stage to assist the human driver.

\begin{remark}
An example of defining $\sigma(t)$ is to assume that the human driver only makes decisions after some time interval $0 < \Delta t \leq T$. Let $K = \lfloor \frac{T}{\Delta t} \rfloor$. Then we have $\sigma(t) = 1$ for $t = k \Delta t$ and $\sigma(t) = 0$ for $t \neq k \Delta t$, $k\in \bbracket{K}$. 
\end{remark}

We use the quantal response (QR) model to capture the human driver's bounded rationality in decision-making. The following example explains the QR model in a static Stackelberg game.

\begin{example}
Assume the follower has a utility matrix $V \in \R^{m\times n}$ in a static Stackelberg game. The QR model finds a mixed strategy $y^* = \arg\max_{y \in \Delta(n)} x^\tp V y - \frac{1}{\lambda} y \log y$ to respond to the leader's committed strategy $x \in \Delta(m)$. This provides a logit choice model on the true utility, i.e., $y^*_i = \frac{\exp(\lambda x^\tp V_i)}{\sum_k \exp(\lambda x^\tp V_k)}$ for every $i \in \{ 1,\dots, n\}$. Here, $V_i$ represents the $i$-th column of the matrix $V$. $\lambda > 0$ is the bounded rationality constant, which can be determined by empirical studies. For comparison, if the follower is rational, she will respond to the leader's committed strategy $x$ with a pure strategy $i^* = \arg\max_i x^\tp V_i$. 
\end{example}

Therefore, we formulate the shared control scheme as a hybrid and regularized dynamic Stackelberg game as follows:
\begin{equation}
\label{eq:dsg}
\begin{split}
    \max_{\by^L} \ & \Exp_{p,\by^L, \by^{F*}} \left[ \sum_{t=0}^{T-1} \gamma^t g^L(x_t, u^L_t, u^{F}_t) + q^L(x_T) \right] \\ 
    \text{s.t.} \ %& x_{t+1} = f(x_t, u^L_t, u^F_t), \quad t = 0,\dots, T-1, \\
    & y^L_{t,x} \in \Delta(m^L), \quad \forall x \in \mX, \ t \in \bbracket{T-1}, \\
    & \by^{F*} = \arg\max_{\by^F} \ \Exp_{p,\by^L, \by^{F}} \left[ \sum_{t=0}^{T-1} \gamma^t g^F(x_t, u^L_t, u^F_t) \right. \\ 
    & \hspace{1cm} + q^L(x_T) \bigg] - \sum_{t=0}^{T-1} \sigma(t) \frac{1}{\lambda} \by^F \log \by^F,  \\ 
    %& \quad \text{s.t.} \quad u^F_t = \varnothing \text{ if } \sigma(t) = 0, \quad t = 0,\dots, T-1. \\
    & \text{s.t.} \ y^F_{t,x} \in \Delta(m^F), \quad \forall x \in \mX, \ t \in \bbracket{T-1}, \\
    & \hspace{5mm} y^F_{t,x} = \be(\varnothing) \text{ if } \sigma(t) = 0, \forall x \in \mX, t \in \bbracket{T-1}. \\
\end{split}
\end{equation}
Here, $\by^L := \{ y^L_{t,x} \in \Delta(m^L)\}_{x \in \mX, t \in \bbracket{T-1}}$ (resp. $\by^F$) is the leader's (resp. follower's) time-state dependent feedback policy trajectory. $q^L(x)$ and $q^F(x)$ are the leader and follower's terminal rewards, which can be specified in advance. $\gamma \in (0,1] $ is the discounted factor. $\be(\varnothing) \in \Delta(m^F)$ is a one-hot vector with all mass on the action $\varnothing$. We use a deterministic transition probability $p(x' \vert x, u^L, u^F) = 1$ if $x' = f(x, u^L, u^F)$ and $0$ otherwise to compute the expectation in \eqref{eq:dsg}.

\subsection{Meta-Learning Problem}
In order to use the underlying game \eqref{eq:dsg} to interact and assist the human driver, the planner requires to know the driver's utility function $g^F$. However, in practice, the planner can only observe the driver's actions. Therefore, the planner needs to learn the human driver's utility function from observations.
Besides, the planner should also be able to work with different drivers and provide as good driving assistance as possible. Each human driver is associated with a type $\theta \in \Theta$ and has a distinguished utility function $g^F_\theta$, which leads to different driving behaviors.
We assume that the total types are finite. Every time a human driver requests to use the vehicle, we assume that the driver's type is drawn from the probability distribution $\mu(\theta)$. It is time-consuming to learn the driver's utility from scratch when the driver is fixed. Our objective is to develop an approach that can quickly generate assistive driving strategies once the type of human driver is identified. Therefore, we formulate a meta-learning problem first to learn a generalized utility function for all types of human drivers. Then, we only use a small amount of data (from driving history) to customize the generalized utility to the driver-specific one and assist driving.

\begin{remark}
In this work, we assume that different types of human drivers share the same decision indicator function $\sigma(t)$, meaning that all drivers make decisions at the same interaction stage over the prediction horizon $T$. While this could be a strong assumption in some cases, it brings extra benefits in strategy analysis and designing the meta-learning algorithm, e.g., parallel implementation. We leave a more general driver's type model as the future work.    
\end{remark}

\section{Stackelberg Meta-Learning} \label{sec:sgmeta}
\subsection{FSE via Dynamic Programming}
Effective learning algorithms require the characterization of the relationship between the equilibrium strategy and the driver's utility $g^F$. We first use dynamic programming (DP) to compute the FSE of the game \eqref{eq:dsg}. We temporarily ignore the type subscript $\theta$ since the FSE structure is the same for all types of drivers. 

Let $V^L_t(x)$ and $V^F_t(x)$ be the leader and follower's value functions for state $x \in \mX$ at time $t$. We have $V^L_T(x) = q^L_T(x)$ and $V^F_T(x) = q^F_T(x)$.
When $\sigma(t) = 0$ for $t \in \bbracket{T-1}$, the follower does not make decisions and takes the action $u^F = \varnothing$. Therefore, the leader updates its value function $V_t$ by solving
\begin{equation}
\label{eq:leader_dp_1}
    V^L_t(x) = \max_{u^L \in \mU^L} g^L(x,u^L, \varnothing) + \gamma \sum_{x' \in \mX} p(x' \vert x, u^L, \varnothing) V^L(x')
\end{equation}
for all $x \in \mX$. The follower  updates its value function according to
\begin{equation}
\label{eq:follower_dp_1}
    V^F_t(x) = g^F(x,u^{L*}_{t,x}, \varnothing) + \gamma \sum_{x' \in \mX} p(x' \vert x, u^{L*}_{t,x}, \varnothing) V^F_{t+1}(x')
\end{equation}
for all $x \in \mX$, where 
\begin{equation}
    u^{L*}_{t,x} = \arg\max_{u^L \in \mU^L} g^L(x,u^L, \varnothing) + \gamma \sum_{x' \in \mX} p(x' \vert x, u^L, \varnothing) V^L(x').
%\begin{split}
%    u^{L*}_{t,x} = \arg\max_{u^L \in \mU^L} &g^L(x,u^L, \varnothing) \\ &+ \gamma \sum_{x' \in \mX} p(x' \vert x, u^L, \varnothing) V^L(x').
%\end{split}
\end{equation}
Thus, $\{ u^{L*}_{t,x}, \varnothing \}_{\forall x \in \mX}$ constitute the FSE at time $t$ when $\sigma(t) = 0$.

When $\sigma(t) = 1$ for $t \in \bbracket{T-1}$, the follower makes decisions and responds to the leader's committed strategy $y^L_{t,x}$ by solving the following regularized problem:
\begin{equation}
\label{eq:follower_dp_2}
    \max_{y^F_{t,x} \in \Delta(m^F)} \ \Exp_{y^F_{t,x}} \left[ g^F(x, u^L, u^F) + \gamma \sum_{x' \in \mX} p(x'\vert x, u^L, u^F)  V^F_{t+1}(x') \right] - \frac{1}{\lambda} y^F_{t,x} \log y^F_{t,x}.
%\begin{split}
%    &\max_{y^F_{t,x} \in \Delta(m^F)} \ \Exp_{y^F_{t,x}} \bigg[ g^F(x, u^L, u^F) \\ 
%    & \ \ + \gamma \sum_{x' \in \mX} p(x'\vert x, u^L, u^F)  V^F_{t+1}(x') \bigg]
%    - \frac{1}{\lambda} y^F_{t,x} \log y^F_{t,x}.
%\end{split}
\end{equation}
To simplify \eqref{eq:follower_dp_2}, we define the composite utility $\tilde{g}^F_{t,x} \in \R^{m^L \times m^F}$ at time $t$ and state $x$ as
\begin{equation}
\label{eq:comp_utility}
\tilde{g}^F_{t,x}(u^L, u^F) = g^F(x, u^L, u^F) + \gamma \sum_{x' \in \mX} p(x'\vert x, u^L, u^F) V^F_{t+1}(x').
%\begin{split}
%    \tilde{g}^F_{t,x}(u^L, u^F) = &g^F(x, u^L, u^F) \\ &+ \gamma \sum_{x' \in \mX} p(x'\vert x, u^L, u^F) V^F_{t+1}(x').
%\end{split}
\end{equation}
Then, \eqref{eq:follower_dp_2} has a closed-form solution according to the QR model:
\begin{equation}
\label{eq:qr_follower}
    y^{F*}_{t,x}(u^F) = \frac{\exp( \lambda \sum_{a} y^L_{t,x}(a) \tilde{g}^F_{t,x}(a, u^F))}{\sum_{b} \exp(\lambda \sum_{a} y^L_{t,x}(a) \tilde{g}^F_{t,x}(a, b))}, \ \forall u^F \in \mU^F,
\end{equation}
where the sums of $a$ and $b$ are taken over the set $\mU^L$ and $\mU^F$, respectively. Using the follower's response \eqref{eq:qr_follower}, the leader updates its value by solving
\begin{equation}
\label{eq:leader_dp_2}
\begin{split}
    V^L_t(x) = 
    \max_{y^L_{t,x} \in \Delta(m^L)} & \ \Exp_{y^L_{t,x}, y^{F*}_{t,x}(y^L_{t,x})} \bigg[ g^L(x, u^L, u^F) \\
    & \hspace{1cm} + \gamma \left. \sum_{x'\in\mX} p(x'\vert x, u^L, u^F) V^L_{t+1}(x') \right]
\end{split}
\end{equation}
for all $x \in \mX$, which can be efficiently solved by gradient methods. Then, $\{ y^{L*}_{t,x}, y^{F*}_{t,x} \}_{x \in \mX}$ constitutes the FSE at time $t$ when $\sigma(t) = 1$, which are time and state dependent probability vectors.

The follower takes $\varnothing$ when $\sigma(t) = 0$, and the FSE does not reveal any information about the follower's utility $g^F$. However, when $\sigma(t) = 1$, the follower's strategy is related to $g^F$ by \eqref{eq:comp_utility}-\eqref{eq:qr_follower}, which we can use to design learning algorithms to estimate $g^F$.

\subsection{Successive Estimation on Driver's Utility} \label{sec:sgmeta.succ}
A human driver samples a pure action $\hat{u}^F_{t,x}$ from her equilibrium strategy $y^{F*}_{t,x}$ at state $x \in \mX$ and time $t \in \bbracket{T-1}$ to drive the vehicle. We encode $\hat{u}^F_{t,x}$ into a one-hot vector $\hat{y}^F_{t,x} \in \Delta(m^F)$ such that $\hat{y}^F_{t,x}( u^F) = 1$ if $u^F = \hat{u}^F_{t,x}$ and $0$ otherwise. We minimize the cross entropy of the observed samples and the driver's mixed strategy to estimate the composite utility $\tilde{g}^F_{t,x}$ in \eqref{eq:qr_follower} and then estimate $g^F$ by \eqref{eq:comp_utility}. 
Assume we have $N$ observed strategy pairs $\mD := \{ [\hat{y}^L_{t,x}]_{(i)}, [\hat{y}^F_{t,x}]_{(i)} \}_{i=1}^N$ for a fixed state $x$ and time $t$, we minimize the following loss to obtain an estimate on $\tilde{g}^F_{t,x}$:
\begin{equation}
\label{eq:cross_entropy}
\begin{split}
    &L(\tilde{g}^F_{t,x}; \mD) = -\frac{1}{N} \sum_{i=1}^N [\hat{y}^F_{t,x}]_{(i)} \log y^{F*}_{t,x} \\
    &= -\frac{1}{N} \sum_{i=1}^N [\hat{y}^F_{t,x}]_{(i)} \log \frac{\exp( \lambda \sum_a [\hat{y}^L_{t,x}]_{(i)}(a) \tilde{g}^F_{t,x}(a, \cdot) )}{\sum_b \exp(\lambda \sum_a [\hat{y}^L_{t,x}]_{(i)}(a) \tilde{g}^F_{t,x}(a, b))}.
\end{split}    
\end{equation}

We note from \eqref{eq:comp_utility} that $g^F$ can be recovered from $\tilde{g}^F_{t,x}$ only if we know the value function $V^F_{t+1}$. Moreover, $V^F_{t+1}$ is affected by future values. Therefore, the estimation is embedded in the DP, and we can use backward propagation to estimate $g^F$ successively.
Also note that minimizing \eqref{eq:cross_entropy} only yields an estimation result at state $x$ and time $t$. However, solving \eqref{eq:cross_entropy} at different states $x$ when $t$ is fixed is independent, which allows us to design parallel learning algorithms on $g^F$.

\subsection{Meta-Learning for Successive Estimation}
The planner can leverage meta-learning to successively estimate a meta utility $g^F$ as a generalized model for all types of human drivers and then adapt the meta utility to fit a specific driver for driving assistance. From the discussion in Sec.~\ref{sec:sgmeta.succ}, meta-learning is conducted at different states and times. We define meta-learning task $\mT(t,x)$ as estimating the driver's composite utility $\tilde{g}^F_{t,x}$ at time $t$ and state $x$. The corresponding learning objective $L_\theta$ is given by \eqref{eq:cross_entropy}. We denote $\mD_\theta$ as the data set for the human driver with type $\theta$ and split them into $\mD^{train}_\theta \cup \mD^{test}_\theta$. The meta-learning task $\mT(t,x)$ is formulated as the following optimization problem:
\begin{equation}
\label{eq:meta}
    \min_{\tilde{g}^F_{t,x}} \quad \Exp_{\theta \sim \mu} [L_\theta(\tilde{g}^F_{t,x} - \alpha\nabla L_\theta(\tilde{g}^F_{t,x}; \mD^{train}_\theta); \mD^{test}_\theta)],
\end{equation}
where $\alpha>0$ is the inner gradient update step size. We use the empirical task distribution to approximate the expectation in \eqref{eq:meta} and obtain
\begin{equation}
    \min_{\tilde{g}^F_{t,x}} \ \frac{1}{|\mT_{batch}|} \sum_{\theta \sim \mu} L_\theta(\tilde{g}^F_{t,x} - \alpha \nabla L_\theta(\tilde{g}^F_{t,x}; \mD^{train}_\theta); \mD^{test}_\theta). 
\end{equation}
Here, $\theta \sim \mu$ is the empirical task distribution of sampled batch tasks $\mT_{batch} = \{ \mT_\theta \}$ from $p$. The inner parameter updates from the meta-parameter $\tilde{g}^F_{t,x,(k)}$:
\begin{equation}
\label{eq:maml_inner}
    \tilde{g}^{F'}_{\theta} = \tilde{g}^F_{x,t,(k)} - \alpha \nabla L_\theta \left( \tilde{g}^F_{t,x,(k)}; \mD^{train}_\theta \right),
\end{equation}
We also use gradient methods to update the meta-objective, and the total update is given by 
\begin{equation}
\label{eq:maml_outer}
\begin{split}
    &\tilde{g}^F_{t,x,(k+1)} = \Tilde{g}^F_{t,x,(k)}  \\ 
    & \quad - \frac{\beta}{\abs{\mT_{batch}}} \sum_{\theta \in \mT_{batch}} \left(I - \alpha \nabla^2 L_\theta \left( \tilde{g}^F_{t,x,(k)}; \mD^{train}_\theta \right) \right) \nabla L_\theta(\tilde{g}^{F'}_\theta; \mD^{test}_\theta),
\end{split}
\end{equation}
where $\beta > 0$ is the meta-learning step size. We summarize the successive meta-learning algorithm in Alg.~\ref{alg:sgmeta.1}, which outputs a meta utility $g^F$. 

\begin{algorithm}[h]
\KwInit decision horizon $T$, $\sigma(t)$, $g^F_{init}$ \;
$k\gets 0$, $g^F_{(k)} \gets g^F_{init}$ \;
\For{$k < $ MAX\_ITER}{
    Sample a batch of tasks $\mT_{batch} \sim \mu$ \;
    Set meta value function $V^L_T(x), V^F_T(x)$, $\forall x \in \mX$ \;
    Sample $\mD^{train}_\theta, \mD^{test}_\theta$, $\theta \in \mT_{batch}$ \;
    \tcc{Perform dynamic programming}
    \For{$t = T-1, \dots 0$}{
        Collect $\mX_t := \bigcup_\theta \mX_{\theta,t}$ from $\mD^{train}_\theta$ at time $t$ \;
        \eIf{$\sigma(t) = 0$}{
            \For{$x \in \mX_t$ (in parallel)}{
                Update $V^L_t(x)$ and $V^F_t(x)$ using $g^F_{(k)}$ based on \eqref{eq:leader_dp_1}-\eqref{eq:follower_dp_1} \;    
            }
        }{
            \For{$x \in \mX_t$ (in parallel)}{
                $\tilde{g}^F_{t,x,(k+1)} \gets$ do meta-learning task $\mT (x,t)$ \;
                $g^F_{(k+1)} \gets$ estimate $g^F$ using $\tilde{g}^F_{t,x,(k+1)}$ and \eqref{eq:comp_utility} \;
                Compute $y^{F*}_t$ \eqref{eq:qr_follower} and update $V^F_t$ \eqref{eq:follower_dp_2} with $g^F_{(k+1)}$ \;
                Solve $y^{L*}_t$ with $g^F_{(k+1)}$ and update $V^L_t$ using \eqref{eq:leader_dp_2} \;
            }
        }
    }
    $g^F_{(k)} \gets g^F_{(k+1)}$ \;
}
$g^F_{meta} \gets g^F_{(k)}$ \;
\KwOut meta utility $g^F_{meta}$ \;
\caption{Successive meta-learning of meta utility.}
\label{alg:sgmeta.1}
\end{algorithm}

\begin{algorithm}
\KwIn state $x$, decision time $t$, batch tasks $\mT_{batch}$, meta utility $g^F_{(k)}$, value function $V^F_{t+1}$ \;
Compute meta composite utility $\tilde{g}^F_{(k)}$ using \eqref{eq:comp_utility} \;
\For{All tasks $\mT_\theta \in \mT_{batch}$}{
    Extract all strategy pairs $\mD'_\theta: = \{\hat{y}^L_{t,x}, \hat{y}^F_{t,x}\}$ from $\mD^{train}_\theta$ \;
    \eIf{$\mD'_\theta$ is empty}{
        $\nabla L_\theta = 0$, $\tilde{g}^F_\theta = \tilde{g}^F_{(k)}$ \;
    }{
        Evaluate $\tilde{g}^{F'}_\theta$ using $\mD'$ and \eqref{eq:maml_inner} \;
    }
    Extract all strategy pairs $\mD''_\theta: = \{\hat{y}^L_{t,x}, \hat{y}^F_{t,x}\}$ from $\mD^{test}_\theta$\;
    \uIf{$\mD''_\theta$ is empty}{
        $\mD''_\theta \gets \mD'_\theta$ \;
    }
}
Update $\tilde{g}^F_{(k+1)}$ using $\tilde{g}^F_\theta, \mD''_{\theta}$ and \eqref{eq:maml_outer} \;
\KwOut meta composite utility $\tilde{g}^F_{(k+1)}$ \;
\caption{Meta-learning $\mT(x,t)$.}
\label{alg:sgmeta.2}
\end{algorithm}

From the assumption that all types of human drivers have the same $\sigma(t)$, meta-learning can be performed in parallel for all $x \in \mX_t$ when $\sigma(t)=1$ because the estimation of $\tilde{g}^F_{t,x}$ does not require information of other states. Besides, we only need to perform meta-learning for the state $x \in \mD^{train}$ instead of all $x \in \mX$. It facilitates the training process.

\begin{remark}
Note that Alg.~\ref{alg:sgmeta.1} does not learn the meta value functions $V^L, V^F$. They are intermediate quantities induced by the meta utility $g^F$, which are used to record information from backward propagation and perform successive estimation on $g^F$.
\end{remark}

\subsubsection{Data Set Structure}
Since we use the FSE as the collaboration plan in driving, for every $\theta \in \Theta$, a data sample in $\mD_\theta$ represents a decision trajectory starting from a certain state $x_0$ at $t=0$. We denote $\mX_{\theta,t}$ as the set of all possible states reached at time $t$. Then, we can represent a data sample as $\{ \hat{y}^L_{t,x}, \hat{y}^F_{t,x} \}_{x \in \mX_{\theta, t}, t \in \bbracket{T-1}}$. The data set $\mD_\theta$ is a collection of decision trajectories starting from different initial states. 
For each data sample, $\hat{y}^F_{t,x}$ is deterministic. It  is either equal to $\be(\varnothing)$ or sampled from $y^{F*}_{t,x}$. The leader's $\hat{y}^L_{t,x}$ can be either stochastic or deterministic.

\subsection{Utility Adaptation and Receding Horizon Planning}
Once a specific driver requests to use the vehicle, the planner quickly adapts the meta utility to the driver-specific utility using a small amount of data and iterations. We summarize the adaptation procedure in Alg.~\ref{alg:sgmeta.3} as follows.

\begin{algorithm}
\KwIn driver type $\theta$, meta utility $g^F_{meta}$ \;
$k \gets 0$, $g^F_{(k)} \gets g^F_{meta}$ \;
\While{$k < C$}{
    Set value function $V^L_T(x), V^F_T(x)$, $\forall x \in \mX$ \;
    Sample $\mD_\theta$ with $\abs{\mD_\theta} = K$ \;
    \For{$t=T-1$}{
        collect $\mX_t$ from $\mD_\theta$ \;
        \eIf{$\sigma(t) = 0$}{
            Update $V^L_t(x), V^F_t(x)$ $\forall x \in \mX_t$, \eqref{eq:leader_dp_1}-\eqref{eq:follower_dp_1} \;
        }{
            \For{$x \in \mX_t$ (in parallel)}{
                Extract $\mD' := \{ \hat{y}^L_{t,x}, \hat{y}^F_{t,x} \}$ from $\mD_\theta$ \; 
                $g^F_{(k+1)} \gets g^F_{(k)} - \alpha \nabla L_\theta(g^F_{(k)}; \mD')$; \tcp{$\nabla L = 0$ if $\mD'$ empty}
            }
        }
    }
    $k \gets k+1$ \;
}
$g^F_\theta \gets g^F_{(k)}$ \;
\KwOut Adapted utility $g^F_\theta$ \;
\caption{Utility adaptation for any specific driver.}
\label{alg:sgmeta.3}
\end{algorithm}

The planner uses the adapted utility function for assistive driving after running Alg.~\ref{alg:sgmeta.3}.
To complete the driving task, we leverage the receding horizon approach to implement the shared control in driving, which is summarized in Alg.~\ref{alg:receding}.

\begin{algorithm}
\KwInit driver type $\theta$, initial state $x_0$ \;
$g^F_\theta \gets$ utility adaptation from Alg.~\ref{alg:sgmeta.3} \;
Set $g^L = g^F_\theta$ for assistive driving \;
\For{$t = 0,1,\dots$}{
    Planner and driver observe the current state $x_t$ \;
    Planner predicts possible states $\mX_{set}$ for $T$ steps starting from $x_t$ \;
    Planner performs DP on $\mX_{set}$ to obtain $y^L_{\tau, x}, y^F_{\tau, x}$, $x \in \mX, \tau \in \bbracket{T-1}$, and announces the driving strategy $y^L_{\tau, x}$ \;
    Planner samples an action $u^L_t \sim y^L_{0, x_t}$ \;
    Driver performs DP to obtain her own strategy $y^F_{\tau, x}$, $x \in \mX, \tau \in \bbracket{T-1}$ \;
    Driver samples an action $u^F_t \sim y^F_{0, x_t}$ \;
    Apply $u^F_{t}, u^F_t$ and obtain $x_{t+1} \gets f(x_t, u^L_t, u^F_t)$ \;
}
\caption{Receding horizon planning for shared control.}
\label{alg:receding}
\end{algorithm}

During the driving, the planner can directly recommend driving actions to the human driver because the planner also estimates the driver's equilibrium strategy (which may not be accurate) when computing the FSE. However, suppose that the human driver takes the recommended action. In this case, the vehicle is, in fact, solely controlled by the planner, and the human driver loses the role of a decision-maker in the shared control framework. It is a degenerate case, and we assume in this work that human drivers are capable of making driving decisions.

\section{Simulations and Evaluations} \label{sec:experiment}

\subsection{Simulation Settings and Data Collection}
We evaluate the developed algorithms with a simulated three-lane driving scenario shown in block 3 of Fig.~\ref{fig:intro}, where the planner and the human driver collaboratively drive the vehicle to the target destination while avoiding obstacles. The horizontal distance (position $p$) belongs to $\{0,\dots, 9\}$ and the lanes (position $y$) are in $\{0,1,2\}$. We assume that the horizontal velocity $v$ has three levels $\{0,1,2\}$, represented by one arrow and double arrows in Fig.~\ref{fig:adapt}. A full state $x = [p,y,v]$. For the planner and driver's action sets, we have $\mU^L = \mU^F = $ \{keep $(\varnothing)$, accelerate (+$v$), decelerate (-$v$), left (+$y$), right (-$y$), stop\}. 
%
\begin{comment}
The driving dynamics are defined as follows:
\begin{gather*}
    p_{t+1} = p_t + v_{t+1}, \\ 
    v_{t+1} = \begin{cases} v_t + a^x_t + b^x_t & o.w. \\ 
    0 & v_t + a^x_t + b^x_t < 0\\ 
    2 & v_t + a^x_t + b^x_t > 2\\
    \end{cases}, \\
    y_{t+1} = \begin{cases} y_t + a^y_t + b^y_t & o.w.\\ 
    0 & y_t + a^y_t + b^y_t < 0 \\ 
    2 & y_t + a^y_t + b^y_t > 2 \\
    \end{cases},
\end{gather*}
where $a^x_t, a^y_t \in \{0,1,-1\}$ indicate that the planner takes the action $\{\varnothing, +v, -v\}$ and $\{\varnothing, +y, -y\}$, respectively. The same interpretation applies to $b^x_t, b^y_t$.    
\end{comment}
%
The horizontal position dynamics is governed by the velocity such that $p_{t+1} = p_{t} + v_{t+1}$; the lane change and the velocity dynamics is directly controlled by the planner and the driver's driving actions.
The target destination is set as $x_{goal} = [9,0,0]$ (reaching $[9,0]$ with zero velocity). We set a terminal reward $q_T(x) = 5$ if $x = x_{goal}$ and 0 otherwise. 
We set the decision horizon $T=5$ for interaction and $\sigma(t) = [1,0,0,1,0]$ to reduce the human driver's cognitive loads in planning. The bounded rationality constant is set as $\lambda = 10$ based on empirical studies \cite{mckelvey1995quantal,pita2010robust}.

We consider five types of human drivers $(\abs{\Theta} = 5)$ with a type distribution $\mu = [0.2, 0.3, 0.1, 0.2, 0.2]$. To generate the driving data, we construct human drivers' ground truth utility based on the following cost: the distance cost $z_1 = c_{11} \abs{p-p_{goal}} + c_{12} \abs{y-y_{goal}}$, the obstacle cost $z_2 = c_{23}\log(c_{21}\abs{p-obs_p} + c_{22} \abs{y-obs_y})$, the collision cost (including driving out of lanes) $z_3 = c_3$, and the turning cost $z_4 = c_4$. The utility is based on the \emph{negative} sum of all four costs. Since the costs are functions of states, we use dynamics to compute the cost for each state-action pair and hence define $g^F(x,u^L, u^F)$ for all $(x, y^L, u^F) \in \mX \times \mU^L \times \mU^F$.
The driver in each type has a different set of parameters. 
Type 1: $c_1 = [0.5, 0.01], c_2 = [0.5,1,1.5], c_3 = 10, c_4 = 0$. 
Type 2: $c_1 = [1,0.1], c_2 = [1,2,1.5], c_3 = 10, c_4 = 0$. 
Type 3: $c_1 = [1.5, 0.1], c_2 = [1.5,2.5,1.5], c_3=10, c_4 = 0$. 
Type 4: $c_1 = [0.5, 0], c_2 = [0.5,0.6,1.5]. c_3 = 10, c_4 = 1$. 
Type 5: $c_1=[0.5, 0.01], c_2=[0.5,0.5,1.5], c_3=10, c_4=1$.
Intuitively, we can label types $2-3$ as aggressive drivers and types $4-5$ as careful drivers because types $2-3$ have zero turning costs and lower obstacle costs when approaching obstacles compared with types $4-5$. Hence, they are less sensitive to obstacle avoidance and changing lanes.

To simulate the driving data, we note from \eqref{eq:qr_follower} that collecting the driver's response data does not require the planner to play its optimal strategy. As long as the planner announces a feasible policy trajectory $\{ \hat{y}^L_{t,x} \}_{x \in \mX, t\in \bbracket{T-1}}$, the follower can respond to it by computing \eqref{eq:follower_dp_1}, \eqref{eq:follower_dp_2}, and \eqref{eq:qr_follower}. Then, we can collect the driver's action data by sampling from $y^{F*}_{t,x}$, $x \in \mX, t \in \bbracket{T-1}$. Therefore, we generate multiple different planner's policy trajectories and apply them to human drivers to collect the driving data.

\subsection{Meta-learning and adaptations results}

\begin{figure}
    \centering
    \begin{subfigure}[t]{0.45\textwidth}
        \centering
        \includegraphics[height=4cm]{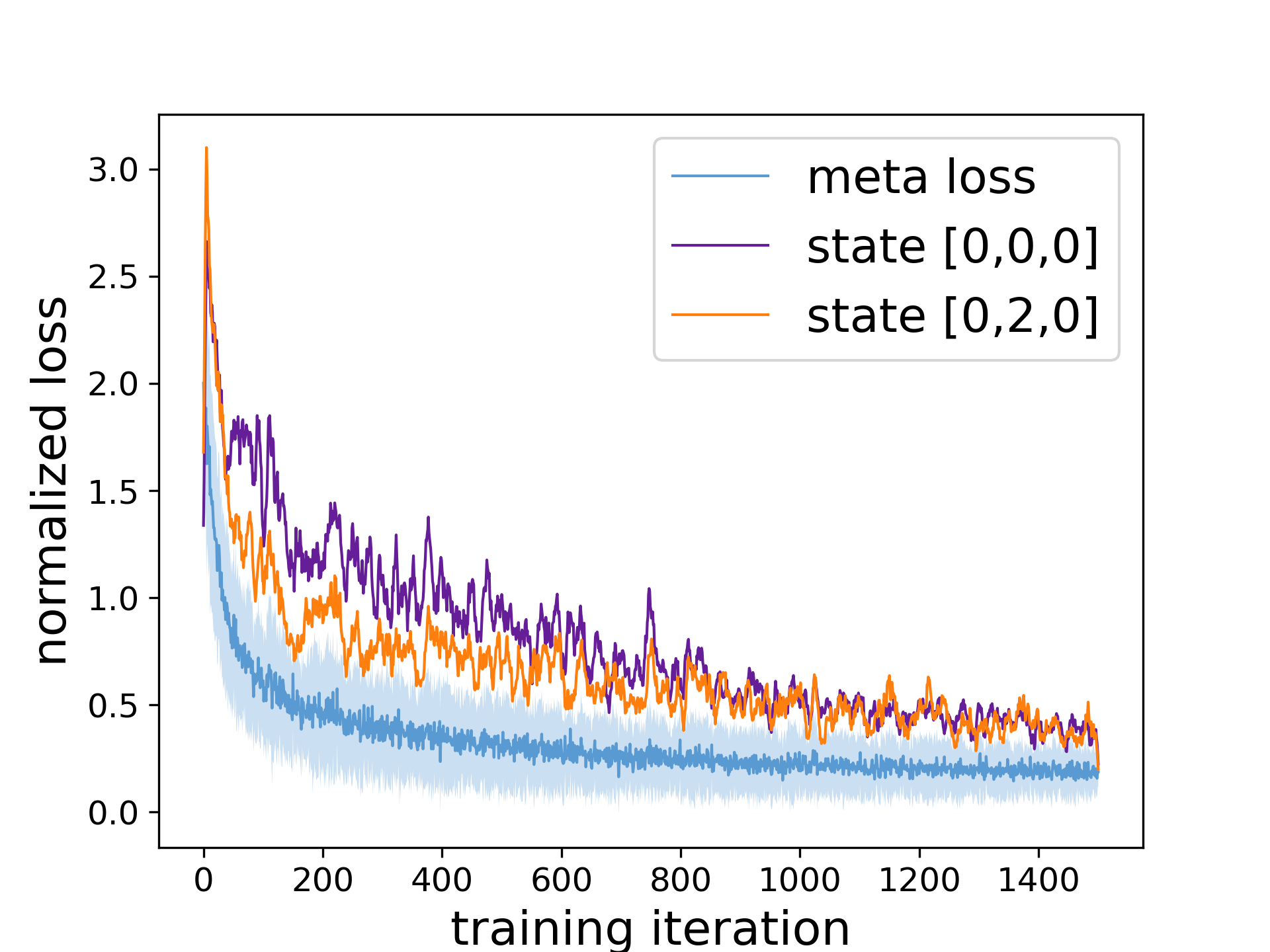}
        %\captionsetup{belowskip=-3pt}
        \caption{Training loss for overall meta-learning and meta-learning at two specific states.}
        \label{fig:meta.1}
    \end{subfigure}
    \hspace{5mm}
    \begin{subfigure}[t]{0.45\textwidth}
        \centering
        \includegraphics[height=4cm]{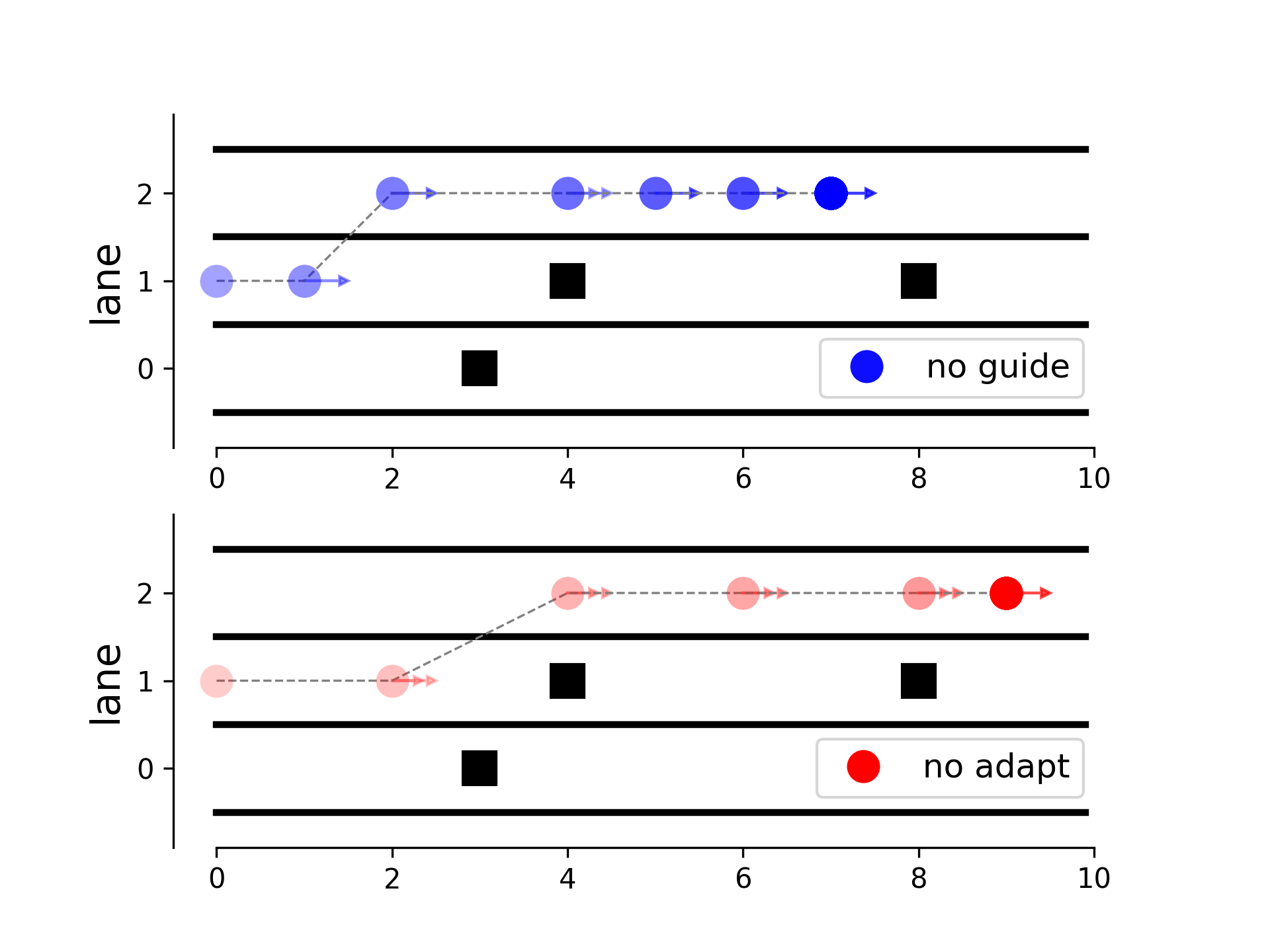}
        %\captionsetup{belowskip=-3pt}
        \caption{[up] Driver-only scheme and [down] assistance with non-adapted model scheme.}
        \label{fig:meta.2}
    \end{subfigure}
    %\captionsetup{belowskip=-12pt}
    \caption{Meta-learning curves and the two failed driving schemes for comparison with adapted results.}
    %\vspace{-2mm}
    \label{fig:meta}
\end{figure}

\begin{figure*}
    \centering
    \begin{subfigure}[t]{0.3\textwidth}
        \centering
        \includegraphics[width=3.5cm]{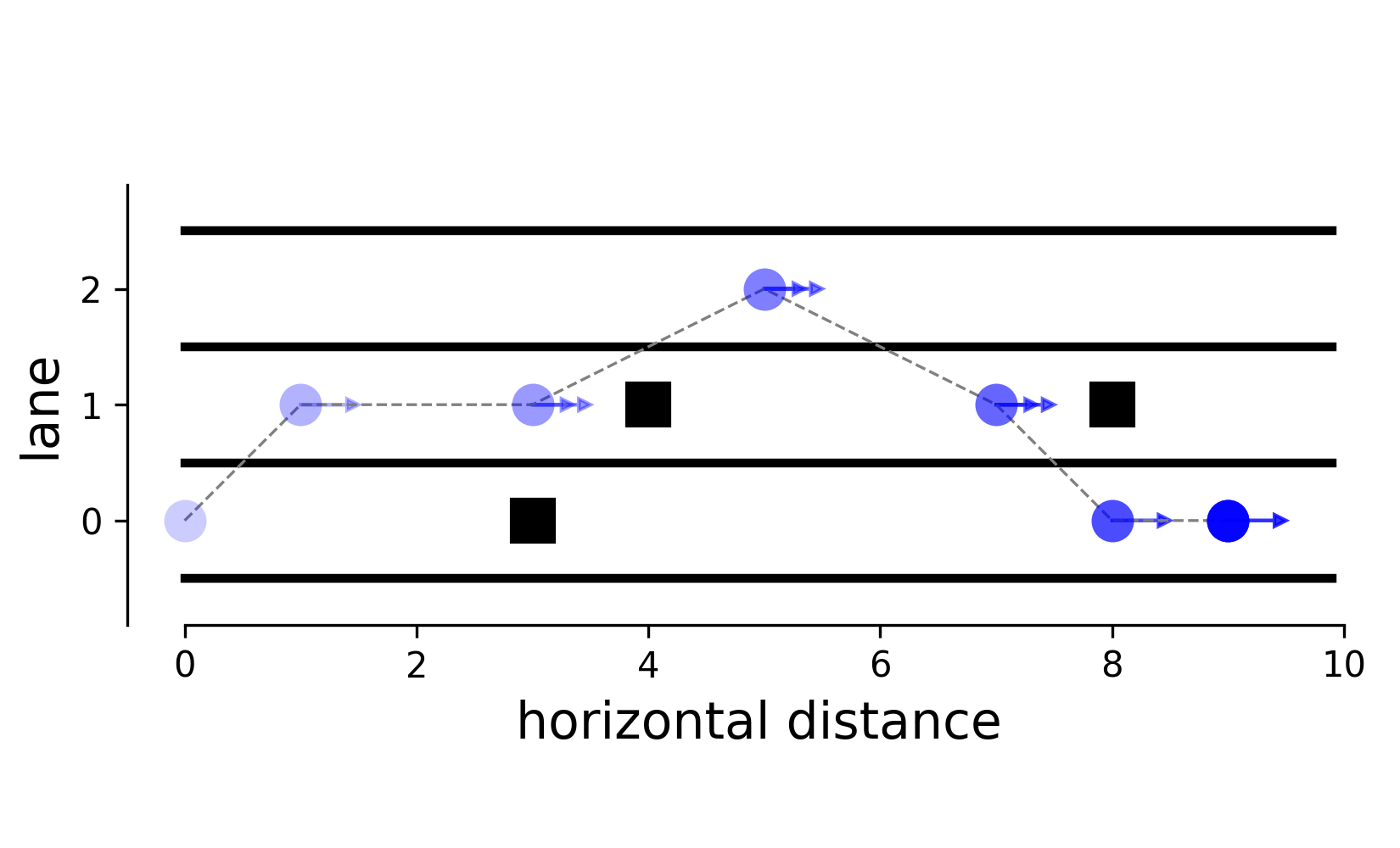}
        %\captionsetup{belowskip=-3pt}
        \caption{Type 1, $x_0=[0,0,0]$.}
        \label{fig:adapt.0}
    \end{subfigure}
    %\hspace{5mm}
    \begin{subfigure}[t]{0.3\textwidth}
        \centering
        \includegraphics[width=3.5cm]{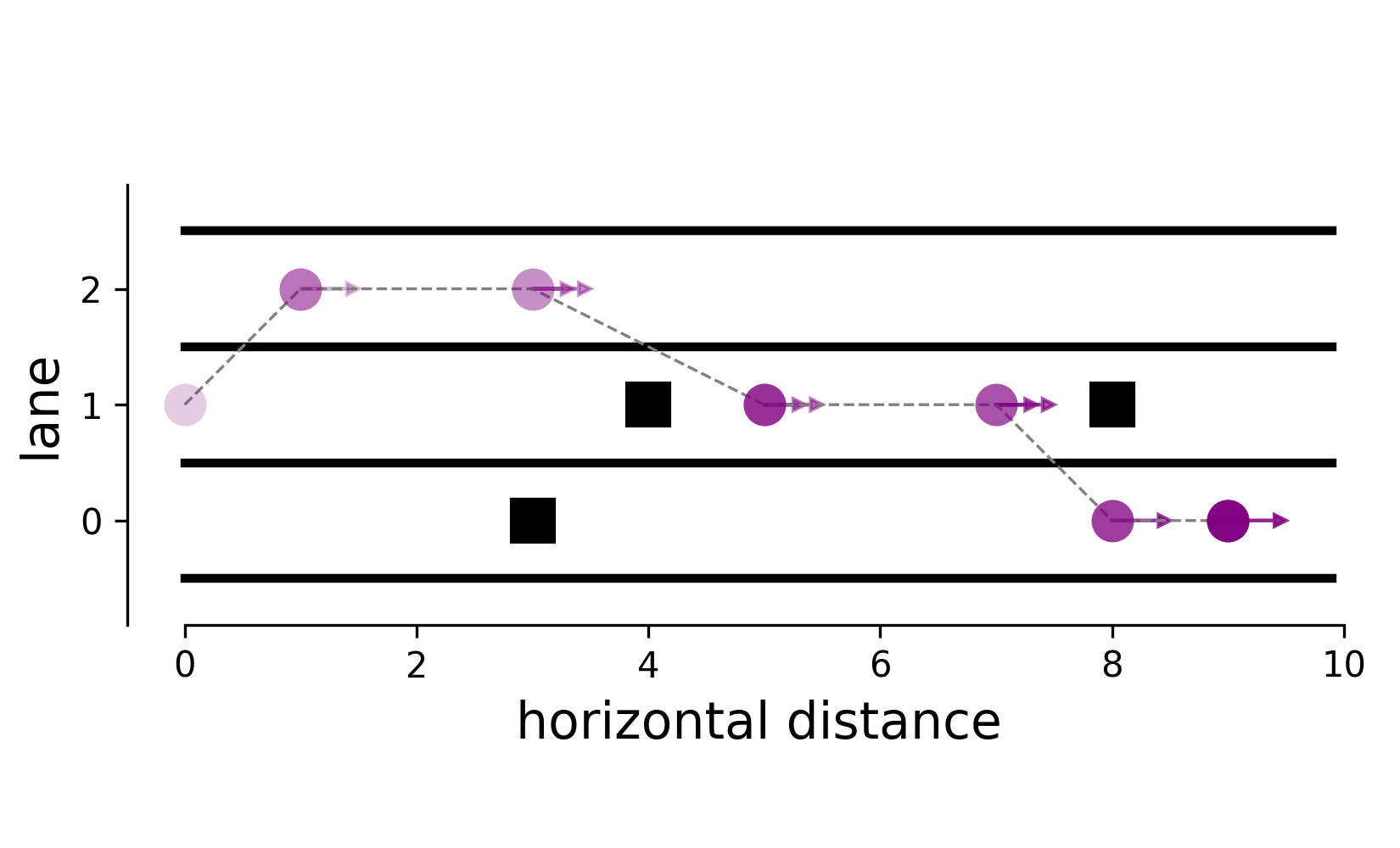}
        %\captionsetup{belowskip=-3pt}
        \caption{Type 2, $x_0=[0,1,0]$.}
        \label{fig:adapt.1}
    \end{subfigure}
    %\hspace{5mm}
    \begin{subfigure}[t]{0.3\textwidth}
        \centering
        \includegraphics[width=3.5cm]{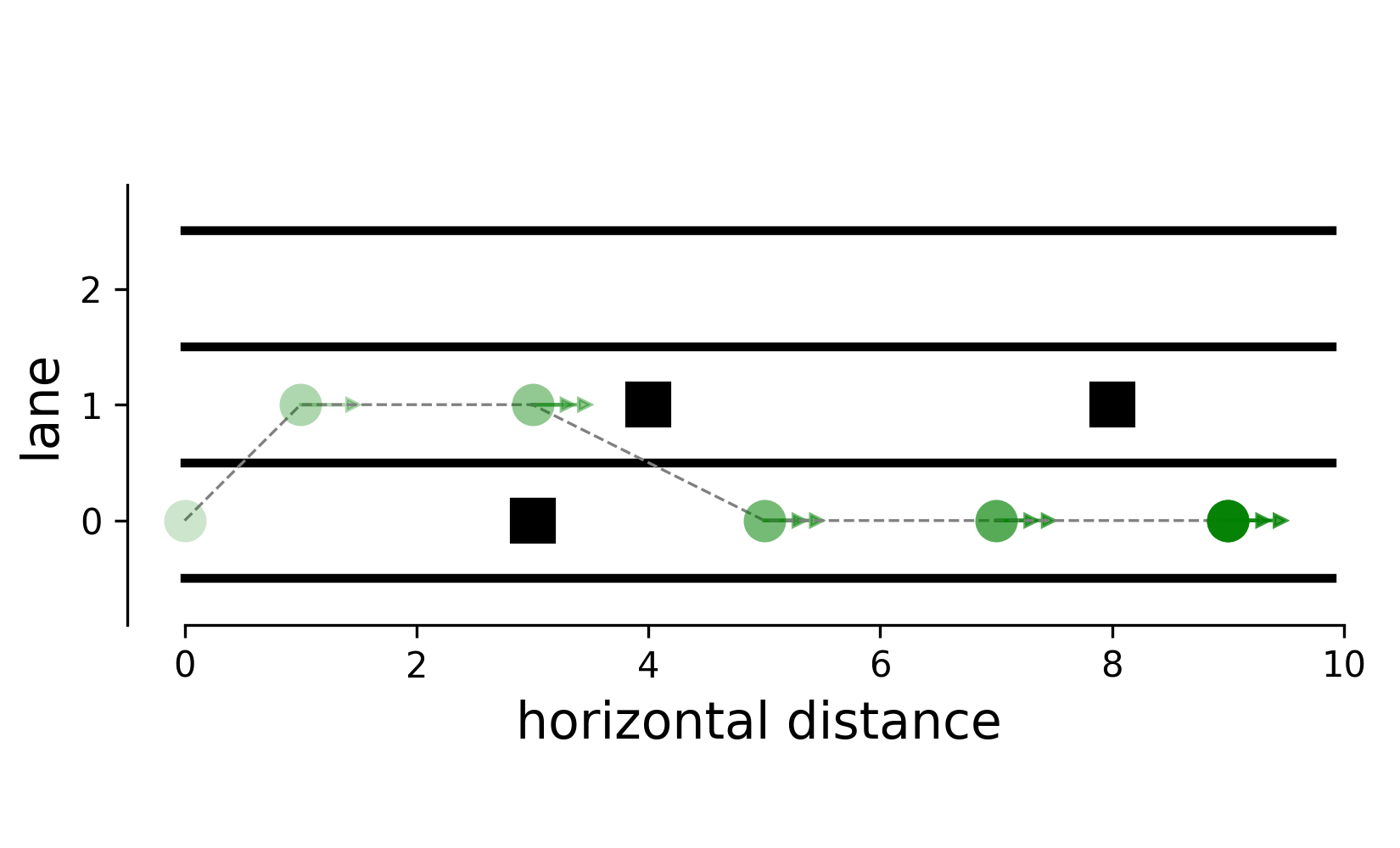}
        %\captionsetup{belowskip=-3pt}
        \caption{Type 3, $x_0=[0,0,0]$.}
        \label{fig:adapt.2}
    \end{subfigure}
    %\hspace{5mm}
    \begin{subfigure}[t]{0.33\textwidth}
        \centering
        \includegraphics[width=3.5cm]{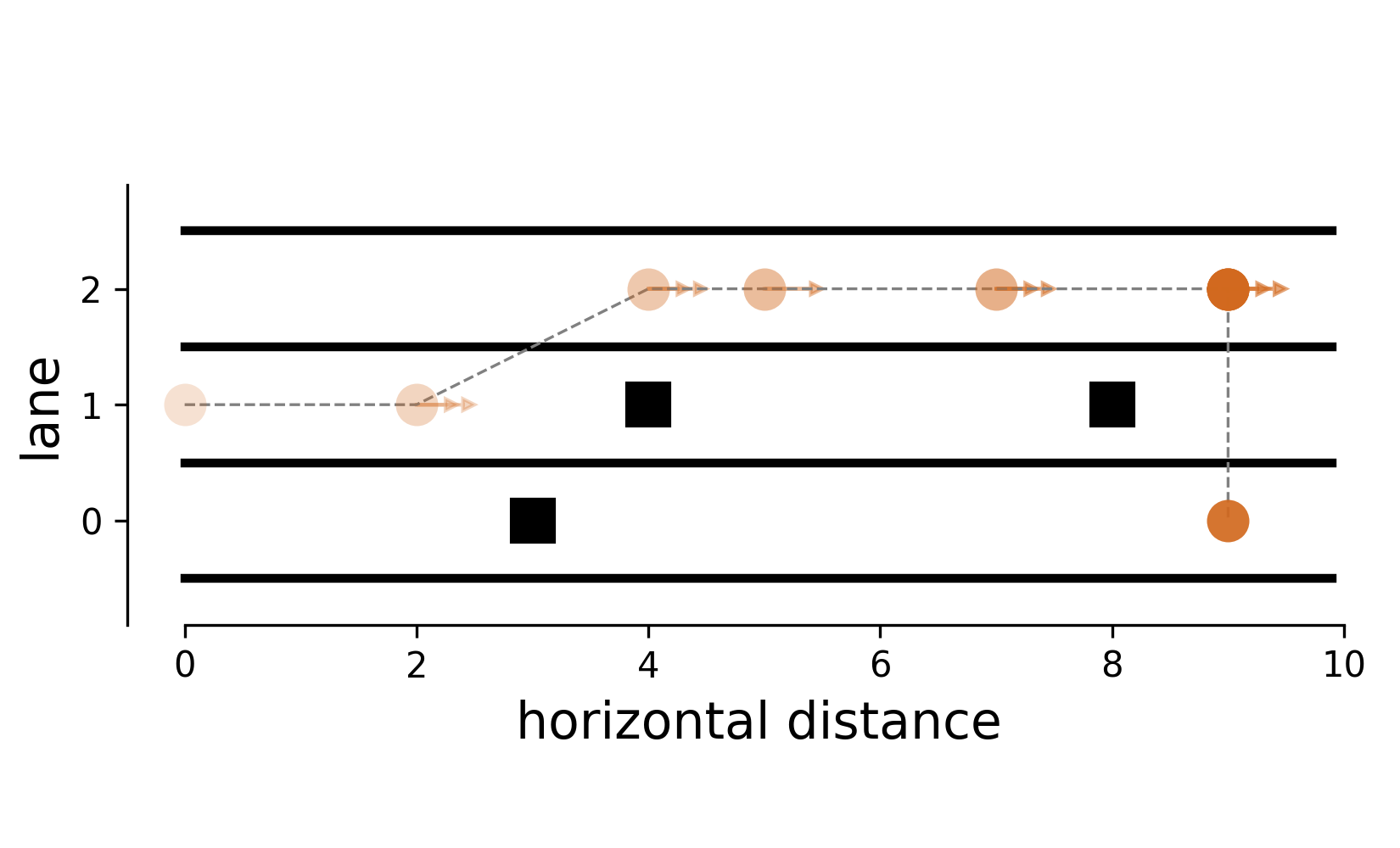}
        %\captionsetup{belowskip=-3pt}
        \caption{Type 4, $x_0=[0,1,0]$.}
        \label{fig:adapt.3}
    \end{subfigure}
    %\hspace{5mm}
    \begin{subfigure}[t]{0.33\textwidth}
        \centering
        \includegraphics[width=3.5cm]{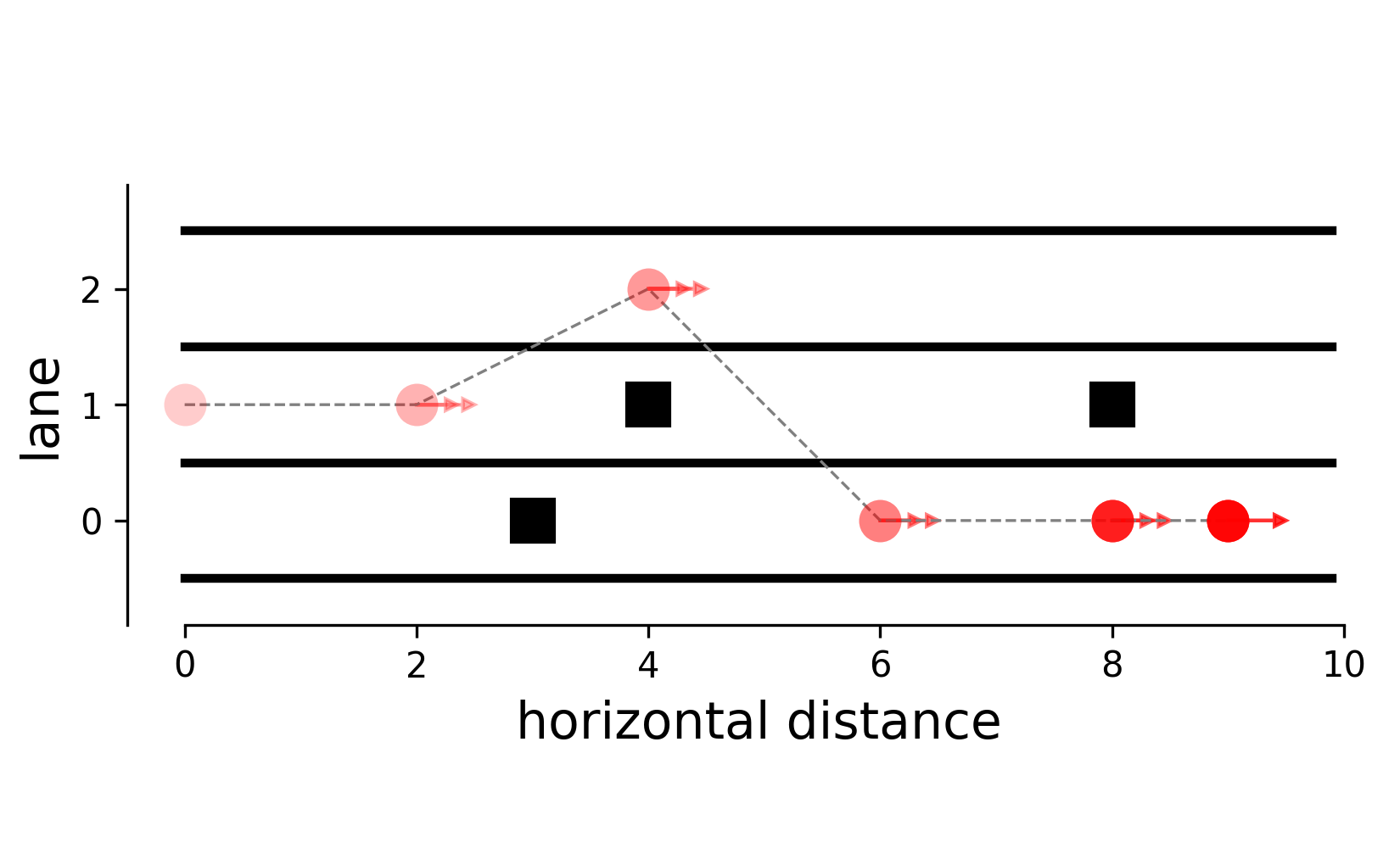}
        %\captionsetup{belowskip=-3pt}
        \caption{Type 5, $x_0=[0,1,0]$.}
        \label{fig:adapt.4}
    \end{subfigure}
    %\captionsetup{belowskip=-12pt}
    \caption{Driving trajectories for different types of human drivers using adapted utilities. The planner successfully assists all human drivers to the target destination, showing the effectiveness of the meta-learning and adaptation algorithms.}
    %\vspace{-2mm}
    \label{fig:adapt}
\end{figure*}

In the meta-learning algorithm Alg.~\ref{alg:sgmeta.1}, we sample $\abs{\mD^{train}_\theta} = 10$ and $\abs{\mD^{test}_\theta} = 5$ in each training iteration, and set the learning step $\alpha = 0.01$ and $\beta = 0.04$. We run 10 simulations to evaluate the learning performance and the learning curve is shown in Fig.~\ref{fig:meta.1}. 
Since meta-learning is performed over different states and time steps, we normalize the overall meta-learning loss by averaging over the state space and prediction horizon and plot it using the blue curve. The overall loss measures how close the learned meta utility is to the true utility $g^F(x,u^L, u^F)$ on average. The decreasing mean value and the decreasing deviation error indicate that the learned meta utility has an increasingly better average performance.
We also plot the mean value of the normalized learning loss (averaged over the decision horizon) of two specific states $x=[0,0,0]$ (purple) and $x=[0,2,0]$ (orange), respectively. They both decrease as learning proceeds. It means that the learned meta utility at these two specific states $g^F_{meta}(x, \cdot, \cdot)$ becomes closer to the averaged ground truth $g^F_\theta(x, \cdot, \cdot)$ of all $\theta \in \Theta$, so that the adaptation to any specific $g^F_\theta(x, \cdot, \cdot)$ becomes more convenient.

We conduct adaptation to obtain the customized utility function for different types of drivers after meta-learning. The planner uses the adapted utility for shared control via Alg.~\ref{alg:receding} for all $\theta \in \Theta$. We set the sampling size $K=10$ and $C=20$ for adaptation. We show the driving trajectories for different types of drivers in Fig.~\ref{fig:adapt}.
As we observe, the planner is able to assist the driver of each type to reach the target destination. For comparison, we experiment with two driving schemes shown in Fig.~\ref{fig:meta.2} to demonstrate the effectiveness of the adaptation results.
We first implement a driver-only scheme where the planner simply takes $\varnothing$ for planning and the driver controls the vehicle. Both the planner and the driver still use Alg.~\ref{alg:receding} to drive. We use the type-5 driver as an example and plot the driving trajectory in the upper half of Fig.~\ref{fig:meta.2}. The driver fails to reach the destination and gets stuck in lane 2. Without assistance, the driver shows difficulty in bypassing the obstacle.
For the second comparison, we make the planner use the non-adapted meta utility to conduct driving assistance. The driving trajectory is shown in the lower half of Fig.~\ref{fig:meta.2}.
Despite the effort of the planner, the vehicle still cannot reach the destination. The meta utility provides an average performance for all types of drivers. It, however,  does not outperform the adapted one for successfully shared control because the adapted utility provides additional information. 

We can also observe that the adapted utility implies the human driver's ground-truth utility. For example, the planner can assist a type-3 driver, who is aggressive and less sensitive to obstacles, in crossing between the obstacles in lane 0 and lane 1 while other human drivers aim to avoid collisions by steering away from the obstacle. Another example is a type-4 driver, who has a negative turning reward and only takes turns in the last move to reach the destination.

\subsection{Uncertainty from Bounded Rationality or Errors}

\begin{figure}
    \centering
    \begin{subfigure}[t]{0.45\textwidth}
        \centering
        \includegraphics[height=4cm]{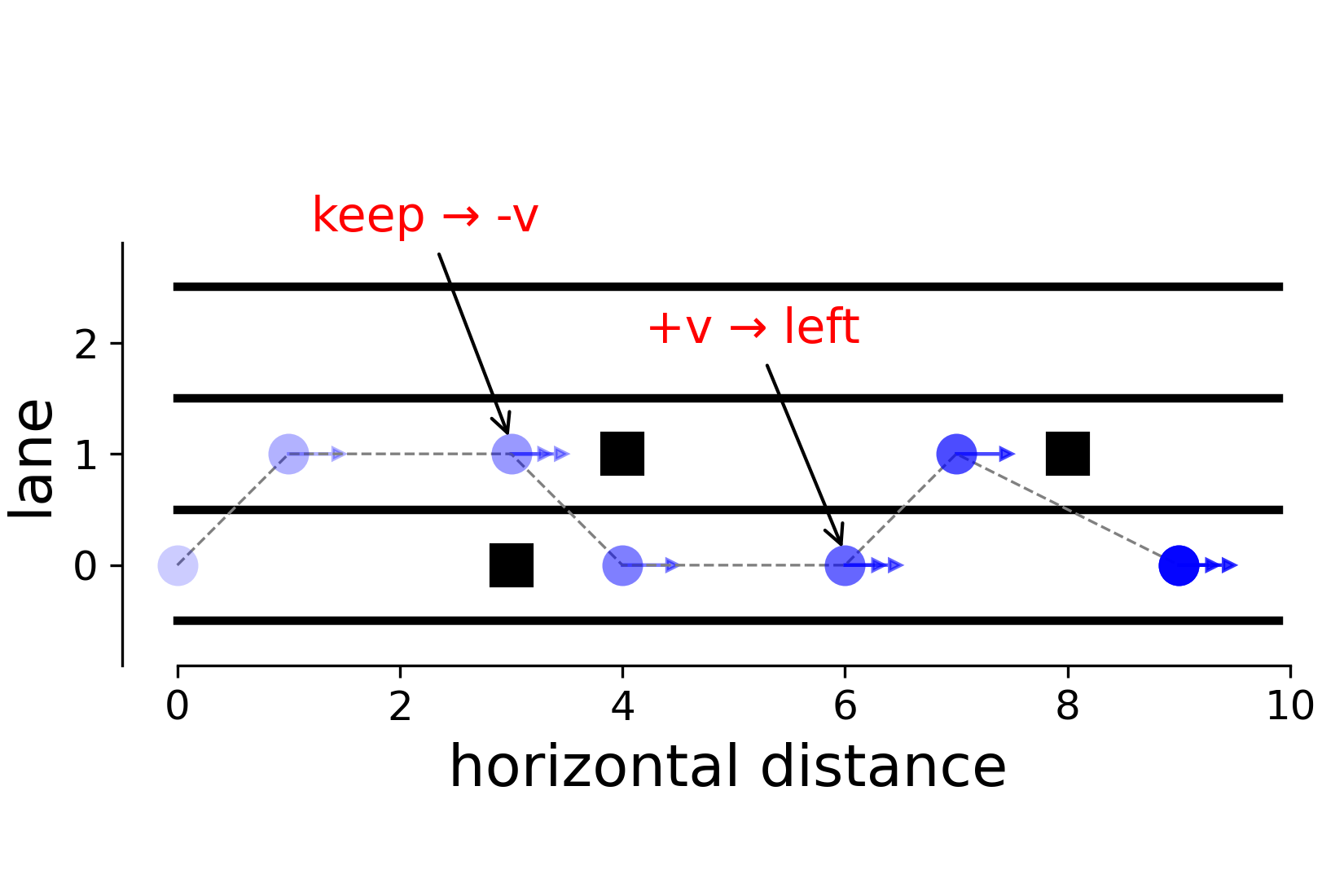}
        %\captionsetup{belowskip=-3pt}
        \caption{Driving trajectory using new driver's actions.}
        \label{fig:robust_type2.1}
    \end{subfigure}
    \hspace{5mm}
    \begin{subfigure}[t]{0.45\textwidth}
        \centering
        \includegraphics[height=4cm]{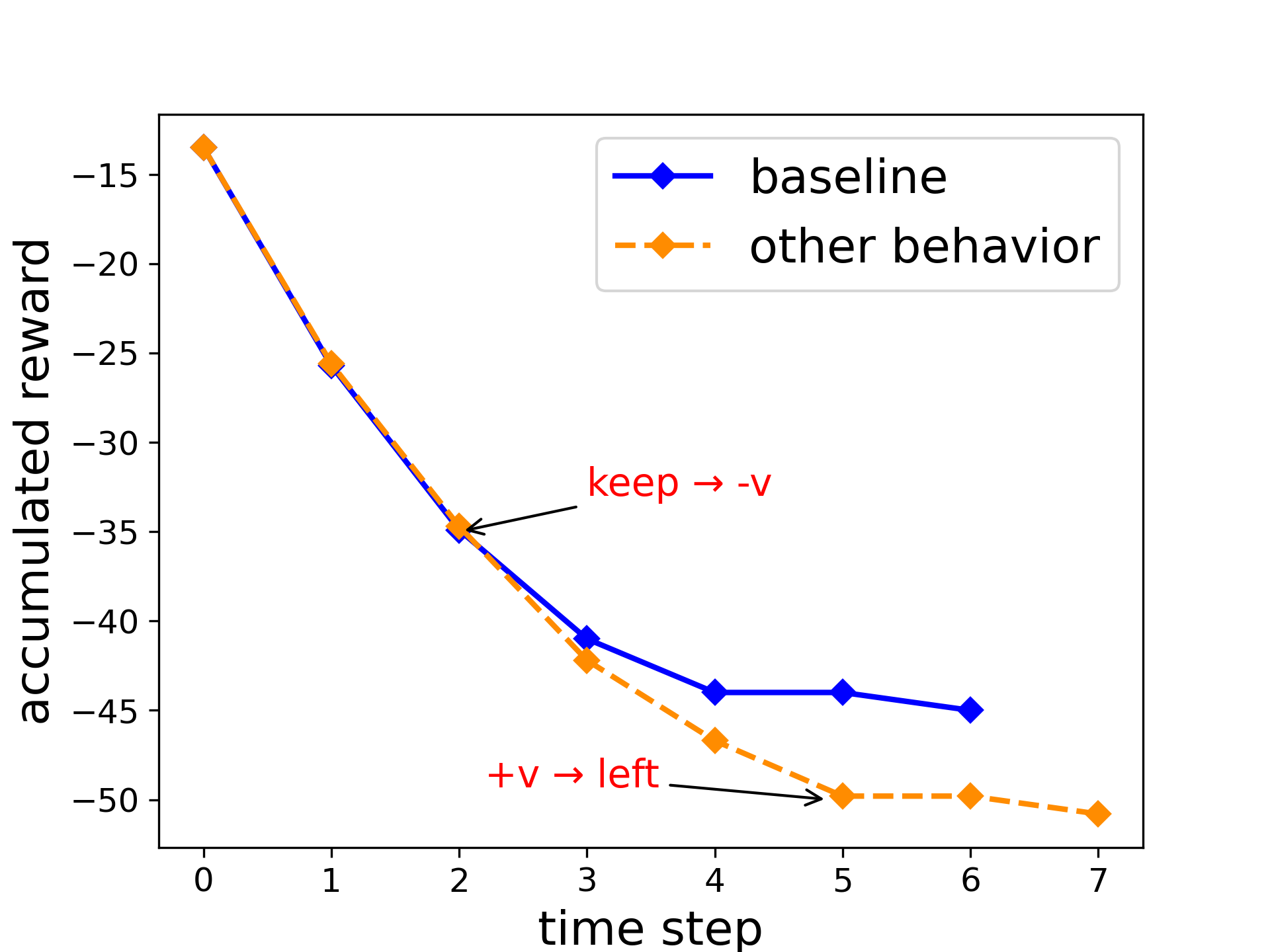}
        %\captionsetup{belowskip=-3pt}
        \caption{Driver's accumulated reward (compared with baseline).}
        \label{fig:robust_type2.2}
    \end{subfigure}
    %\captionsetup{belowskip=-12pt}
    \caption{Simulation results for the type-3 driver with new sampled and perturbed actions compared with the baseline in Fig.~\ref{fig:adapt.2}.}
    %\vspace{-2mm}
    \label{fig:robust_type2}
\end{figure}

\begin{figure}
    \centering
    \begin{subfigure}[t]{0.45\textwidth}
        \centering
        \includegraphics[height=4cm]{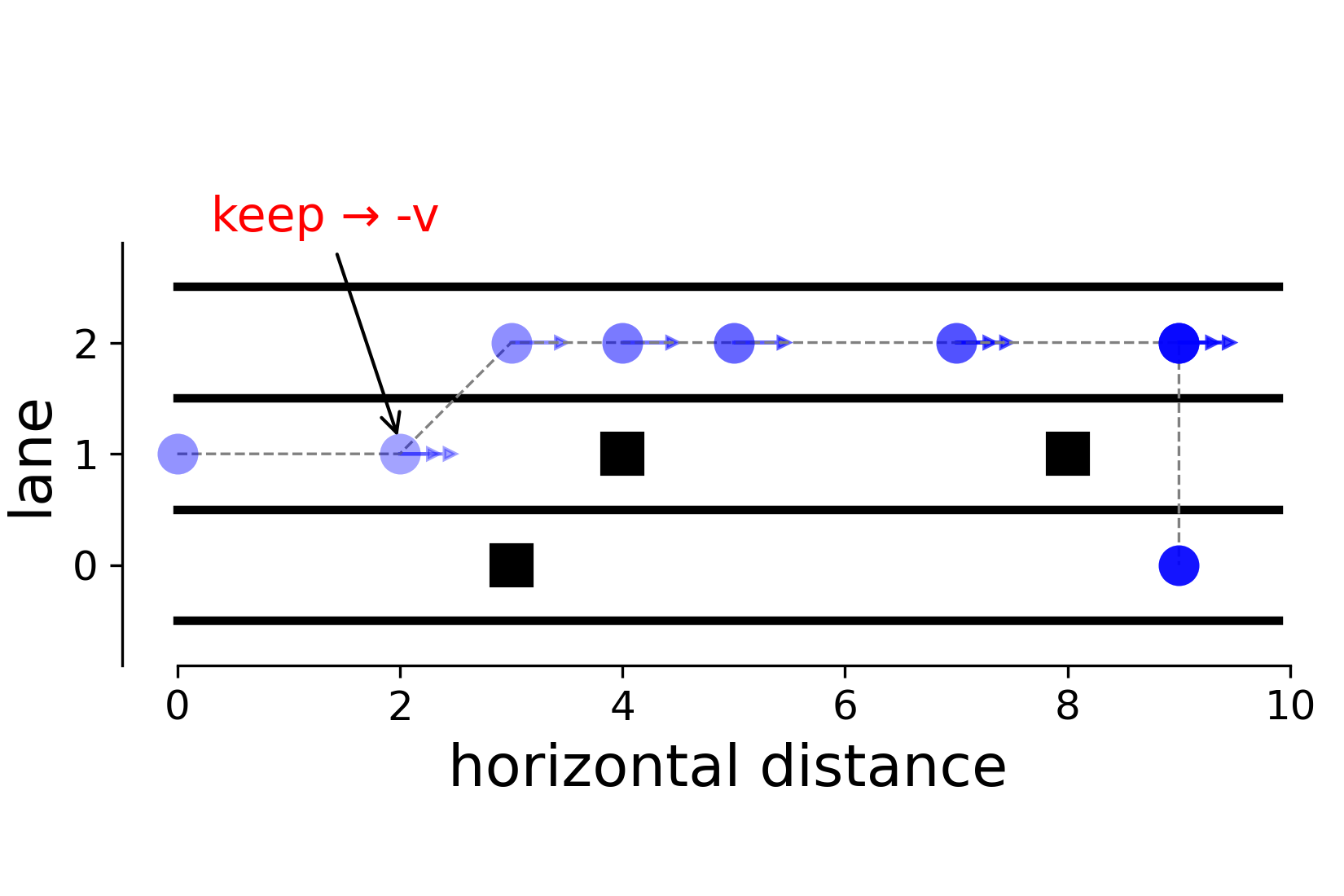}
        %\captionsetup{belowskip=-3pt}
        \caption{Driving trajectory using new driver's actions.}
        \label{fig:robust_type4.1}
    \end{subfigure}
    \hspace{5mm}
    \begin{subfigure}[t]{0.45\textwidth}
        \centering
        \includegraphics[height=4cm]{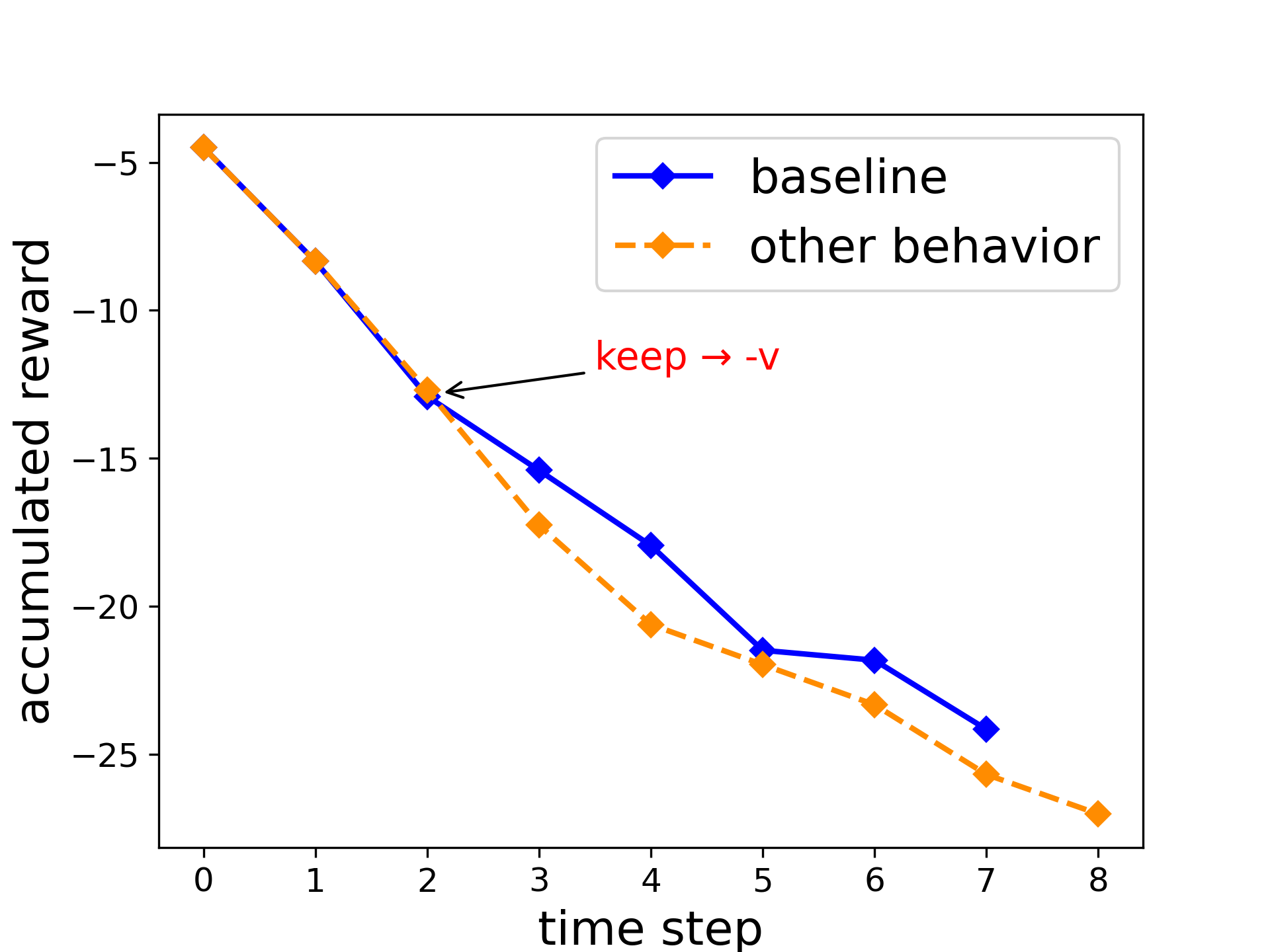}
        %\captionsetup{belowskip=-3pt}
        \caption{Driver's accumulated reward (compared with baseline).}
        \label{fig:robust_type4.2}
    \end{subfigure}
    %\captionsetup{belowskip=-12pt}
    \caption{Simulation results for the type-5 driver with new sampled action compared with the baseline in Fig.~\ref{fig:adapt.4}.}
    %\vspace{-2mm}
    \label{fig:robust_type4}
\end{figure}

Human drivers bring extra uncertainties in driving either because of bounded rationality or driving errors. We show that our learned utility and the shared control framework are robust to these uncertainties. 
We examine type-$3$ and type-$5$ human drivers and use the adapted results in Fig.~\ref{fig:adapt} as the baseline.
For the type-$3$ human driver, when she is in the state $x = [3,2,2]$, she has a probabilistic choice on the action ``keep $(\varnothing)$" or ``decelerate $(-v)$" since she is close to the obstacle, but the probability of selecting the former is greater than the latter. The baseline trajectory in Fig.~\ref{fig:adapt.2} selects ``keep $(\varnothing)$". In the next numerical experiments, we assume the human driver selects ``decelerate $(-v)$" due to action sampling, and mis-selects ``left $(+y)$" instead of ``accelerate $(+v)$" at the state $x = [6,0,2]$. We plot the accumulated reward and the driving trajectory in Fig.~\ref{fig:robust_type2}. 
Since we use negative rewards, the accumulated reward is better if close to $0$. As we observe in Fig.~\ref{fig:robust_type2.2}, the accumulated reward becomes worse when the follower selects ``decelerate" $(-v)$ at state $x=[3,2,2]$. However, the shared control framework can still assist the human driver in bypassing the obstacles in lane $0$ and lane $1$, but with a slower velocity. When the human driver mis-selects the action and turns the vehicle to lane 1, the planner identifies the situation and assists the driver in getting back to lane 0 to reach the target destination. From Fig.~\ref{fig:robust_type2.1}, the planner with the learned utility still achieves successful driving assistance, although it is one step later than the baseline.

For the type-$5$ human driver who is more sensitive to obstacles, she has a positive probability of selecting ``keep $(\varnothing)$" and ``decelerate $(-v)$" at the state $x=[2,1,2]$ but the former one has a larger probability. The baseline trajectory in Fig.~\ref{fig:adapt.4} selects ``keep $(\varnothing)$" to drive. In the next simulation, we assume that the human driver selects ``decelerate $(-v)$". The accumulated reward in Fig.~\ref{fig:robust_type4.2} starts to deviate from the baseline when the new action is taken. The vehicle still approaches the destination in lane 2 and finally reaches the target destination. The reason for staying in lane 2 can be explained by the turning cost. The planner still achieves successful driving with only one step later than the baseline.

\section{Conclusion} \label{sec:conclusion}
Shared control facilitates a seamless and comfortable transition between human-led and autonomous driving. We have introduced a Stackelberg meta-learning framework to design driver-vehicle shared control, which improves the efficiency and safety of human-robot teaming. 
Our framework captures the asymmetric human-vehicle interactions in driving using a dynamic Stackelberg game. It also characterizes uncertainties and cognitive loads in human decision-making through the quantal response model and the decision indicator function.
By using a successive estimation approach, we develop meta-learning and adaptation algorithms that enable the ADAS to design fast and effective driving strategies to collaborate with diverse human drivers. These algorithms have demonstrated robustness to human errors and probabilistic selection of driving actions by a lane-changing collision avoidance driving problem. 

Our future endeavors involve generalizing the framework to fit more realistic environments for comprehensive evaluation. We would also extend our human-vehicle teaming framework to enable the ADAS to collaborate with human drivers possessing different decision indicator functions.


\bibliographystyle{splncs04}
\bibliography{mybib}
\end{document}